\documentclass[runningheads]{llncs}

\usepackage{eccv}

\usepackage{eccvabbrv}

\usepackage{graphicx}
\usepackage{booktabs}

\usepackage[accsupp]{axessibility}  % Improves PDF readability for those with disabilities.

\usepackage{hyperref}

\usepackage{orcidlink}

\definecolor{cvprblue}{rgb}{0.21,0.49,0.74}
\usepackage{mathrsfs}
\usepackage{tcolorbox}
\usepackage{pifont}
\usepackage[table]{xcolor}
\usepackage{multirow}
\usepackage{colortbl}
\usepackage{comment}

\newcommand{\cmark}{\ding{51}}  % ✓
\newcommand{\xmark}{\ding{55}}  % ✗

\definecolor{mycolor}{RGB}{20, 0, 145}  
\newtcolorbox{mycolorbox}[1][]{colframe=mycolor, colback=mycolor!4!white, title=#1}

\definecolor{CHATC}{HTML}{0077E5}
\definecolor{CHATH}{HTML}{0095D8}
\definecolor{CHATA}{HTML}{0062E6}
\definecolor{CHATT}{HTML}{0067BC}

\newcommand{\chat}{
  \textbf{
    \textcolor{CHATC}{C\kern-0.3em}
    \textcolor{CHATH}{H\kern-0.35em}
    \textcolor{CHATA}{A\kern-0.35em}
    \textcolor{CHATT}{T\kern-0.35em}
    \textcolor{Black}{:}
  }
}

\begin{document}

\title{Conversational Human Audio-visual Talking Dialogue Generation} 


\author{Junhao Song\inst{1}\orcidlink{0009-0009-5178-0258} \and
Lluis Guasch\inst{2}\orcidlink{0000-0002-0526-0640} \and
Xilin He\inst{3}\orcidlink{0009-0003-3858-4179} \and 
Zhongyu Yang\inst{4}\orcidlink{0009-0003-0190-8256} \and
Yingfang Yuan\inst{5}\orcidlink{0000-0002-8925-9267} \and
Weicheng Xie\inst{6}\orcidlink{0000-0001-8946-7472} \and
Linlin Shen\inst{7,8}\orcidlink{0000-0003-1420-0815} \and
Haijun Lin\inst{9}\orcidlink{0000-0002-0743-0762} \and
Shizhe Liu\inst{10}\orcidlink{0009-0000-0283-6803} \and
Wei Pang\inst{4}\orcidlink{0000-0002-1761-6659} \and
Siyang Song$^\dagger$\inst{11}\orcidlink{0000-0003-2339-5685}
}

\authorrunning{J.~Song et al.}

\institute{Department of Computing, Imperial College London, UK \and
Department of Earth Science \& Engineering, Imperial College London, UK \and
Mohamed bin Zayed University of Artificial Intelligence, UAE \and
Department of Computer Science, Heriot-Watt University, UK \and
School of Computer Science, Northumbria University, UK \and
College of Computer Science \& Software Engineering, Shenzhen University, China \and
School of Artificial Intelligence, Shenzhen University, China \and
Guangdong Provincial Key Laboratory of Intelligent Information Processing, Shenzhen University, China \and
School of Engineering and Design, Hunan Normal University, China \and
Department of Computer Science, University of Oxford, UK \and
Department of Computer Science, University of Exeter, UK\\
\email{junhao.song23@imperial.ac.uk} and \email{s.song@exeter.ac.uk} \\
$\dagger$ Corresponding author
}

\maketitle

\begin{abstract}
  Large-scale dyadic interactive audio-visual dialogue (DIAD) datasets provide fundamental data resources for developing humanoid interactive virtual agents and digital humans. However, collecting such data is time-consuming, expensive, and ethically sensitive. To address this, we propose CHAT, a new dyadic interactive audio-visual dialogue generation (DIADG) framework that generates diverse, paired, and mutually responsive speech-face dialogue clips from a single textual prompt. CHAT unifies large language models and talking face models with interactive audio and facial behaviour refinement modules, enabling the generation of aligned dyadic dialogue clips with diverse contents and facial identities. Experiments show that CHAT outperforms existing related methods designed for similar tasks under both objective and subjective evaluations. Moreover, our synthesised CHAT-AVD-50k dataset serves as effective pre-training data for downstream interactive head generation, consistently improving PerFRDiff and ReactDiff on REACT 2024. CHAT offers a scalable alternative to the costly and ethically sensitive collection of real dyadic interaction data.
\end{abstract}

%%%%%%%%%%%%%%%%%%%%%%%%%%%%%%%%%%%%%%%%%%%%
%%%             Introduction            %%%%
%%%%%%%%%%%%%%%%%%%%%%%%%%%%%%%%%%%%%%%%%%%%
% \vspace{-3mm}
\section{Introduction}
\label{sec:intro}

\textbf{D}yadic \textbf{I}nteractive \textbf{A}udio-visual \textbf{D}ialogue (DIAD) refers to paired audio-visual recordings capturing the speech, facial behaviours, and mutual responsiveness of two individuals engaged in natural dyadic interactions. Large-scale DIAD datasets are essential for training and evaluating intelligent human-computer interaction (HCI) systems, including virtual agents \cite{wang2024tutor} and digital companions \cite{kovavcevic2024personality}, which must model not only verbal content but also non-verbal cues and interpersonal dynamics. However, collecting large-scale DIAD datasets remains challenging due to privacy and ethical constraints, demographic imbalance, annotation cost, and the complexity of synchronising multi-modal signals in natural dyadic settings \cite{kim2023dcface}. As a result, existing DIAD datasets possess limited size, interaction scenario diversity, and demographic coverage. Motivated by this, we argue that scalable synthesis of large-scale, demographically diverse DIAD datasets containing paired interactive audio-visual-text dialogues from anonymised textual prompts is an effective alternative solution.

\begin{figure}[t]
    \centering
    \includegraphics[width=\linewidth]{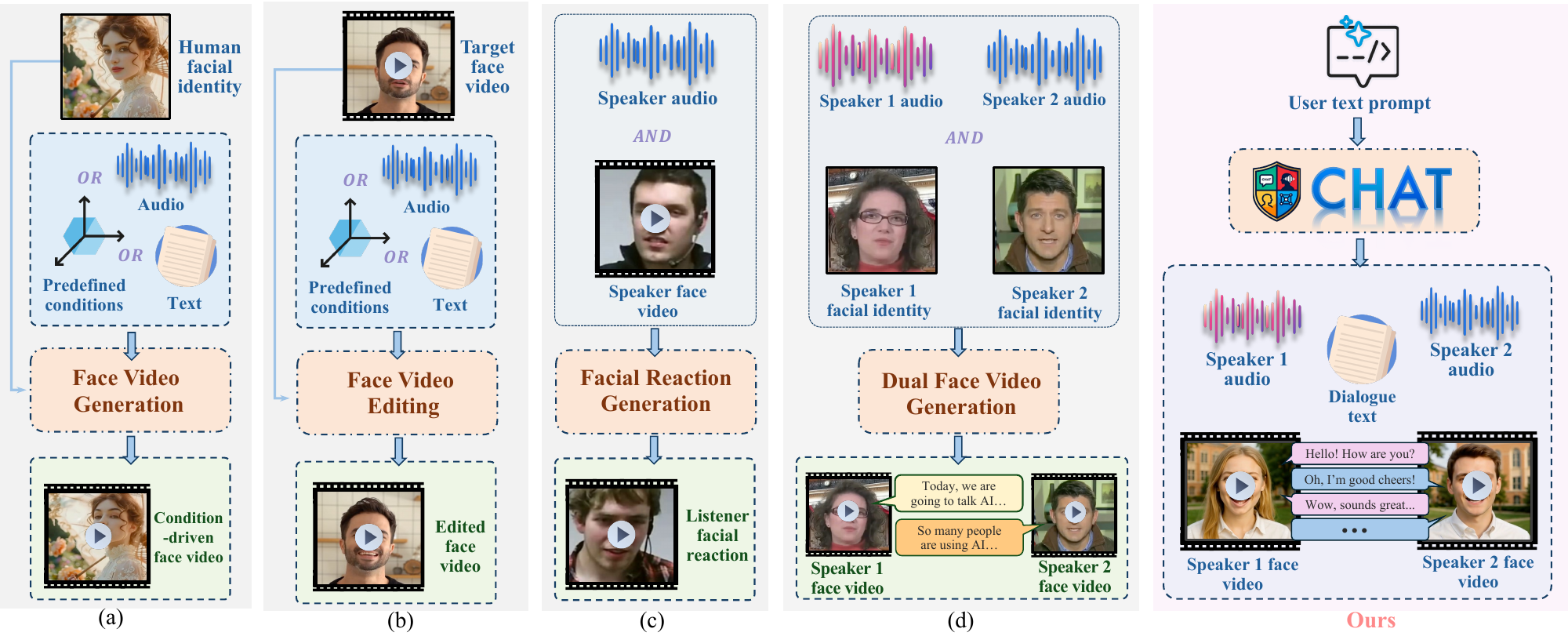}
    \caption{\textbf{Existing methods:} \textbf{(a) Condition-driven face video generation} synthesises a \textit{non-interactive face video} based on pre-defined conditions (e.g., text/audio) and a facial identity; \textbf{(b) Face video editing} utilises pre-defined conditions to modify the given \textit{non-interactive face video}; \textbf{(c) Facial reaction generation} takes human audio-visual behaviours to generate the conversational partner's \textit{interactive non-verbal} facial reaction video; and \textbf{(d) Interactive talking face video pair generation} takes a pair of pre-defined conditions to accordingly generate \textit{a pair of verbally interactive face videos}. \textbf{Our CHAT} enables the generation of diverse, \textit{verbally and non-verbally interactive dyadic audio-visual conversations} directly from a single text prompt, without requiring pre-defined video or audio inputs.}
    \label{fig:teaser}
\end{figure}

%%% Paragraph 2
While the expected DIAD clip pair is to contain two audio-visual clips containing interactive speeches and facial behaviours expressed by a pair of individuals, a large part of existing facial behaviour synthesis methods \cite{xu2024vasa,cui2025hallo3,chen2025fasttalker} can only generate independent and non-interactive face videos from a textual/audio sentence \cite{jang2024faces,choi2024text,cui2025hallo4} or modify a pre-defined face video \cite{pang2023dpe,li2024latentsync,deng2025degstalk} (\textbf{Limitation 1}). Although some advanced approaches \cite{yan2024dialoguenerf,ng2024audio,zhu2025infp,chatziagapi2025av} can already generate paired interactive audio-visual clips, they are still driven by pre-defined conversational speeches/transcripts, i.e., they are not only unable to generate interactive verbal behaviours on their own but also unable to account for how one individual's behaviours are influenced by the other's non-verbal vocal and facial behaviours. Although a few recently proposed facial reaction generation (FRG) methods \cite{xu2026reversible,zhu2024perfrdiff,luo2024reactface,song2025react,song2022learning} can generate facial behaviour videos that are responsive to the input conversational partner's non-verbal behaviours, they are still unable to produce paired human-human interaction audio-visual clips (\textbf{Limitation 2}).

%%% Paragraph 3 
Alternatively, recent large language models (LLMs) \cite{dubey2024llama,guo2025deepseek,yang2025qwen3,comanici2025gemini} enable the generation of diverse human-human dialogues. Although most of them are limited to outputting only diverse appropriate interactive textual dialogues from a user-defined textual prompt or chat contents \cite{liang2022emotional}, a few advanced LLMs \cite{pmlrkondratyuk24a,liu2025ola} can synthesise audio-visual clips. However, to the best of our knowledge, none of the existing LLMs can directly generate diverse, realistic, synchronised, and mutually responsive human-human DIAD clips (\textbf{Limitation 3}).

%%% Paragraph 4
Given the importance of DIADs in developing intelligent HCI systems, this paper introduces a new task, called \textbf{Dyadic Interactive Audio-visual Dialogue Generation (DIADG)}, which requires the developed model to automatically generate a set of diverse but contextually appropriate audio-visual dialogues from a single user-defined anonymous textual prompt, where each dialogue contains interactive speech and facial behaviours expressed by a pair of individuals engaged in a human-human dyadic interaction. In this sense, we propose a DIADG framework, called \textbf{C}onversational \textbf{H}uman \textbf{A}udio-visual \textbf{T}alking (\textbf{CHAT}) Dialogue Generation that consists of three key modules: (i) a \textbf{Textual Dialogue Generation (TDG)} module for creating a set of diverse and fine-grained textual dyadic dialogues; (ii) a \textbf{Dyadic Audio Dialogue Generation (DADG)} module for generating paired, emotion-aware speeches aligned with the corresponding textual dialogues; and (iii) an \textbf{Interactive Facial Behaviour Generation (IFBG)} module for generating mutually responsive and temporally coherent face video pairs aligned with the corresponding interactive speech and textual dialogues. Fig. \ref{fig:teaser} compares our CHAT with existing related methods discussed above. The main contributions of this paper are summarised as follows:
\begin{itemize}
    \item We formally define a new human-centred AI task called DIADG, and propose a DIADG framework called CHAT which can generate a set of diverse audio-visual DIAD clips from a single customisable textual prompt, facilitating scalable, demographically diverse DIAD dataset creation without manual data collection.

    \item We introduce an Interactive Audio Refinement (IAR) block together with a novel trainable Interactive Facial Behaviour Refinement (IFBR) block, transforming the non-interactive, unemotional speeches and face videos produced by pre-trained models into emotional, synchronised and mutually responsive DIAD clips.

    \item Experiments show that CHAT not only outperforms related methods in synchronisation and interactivity while achieving competitive visual quality, but also provides effective pre-training data for downstream Interactive Head Generation (IHG), consistently improving models over real-data-trained counterparts on the REACT benchmark \cite{song2024react}.
\end{itemize}

%%%%%%%%%%%%%%%%%%%%%%%%%%%%%%%%%%%%%%%%%%%%
%%%             Related Work            %%%%
%%%%%%%%%%%%%%%%%%%%%%%%%%%%%%%%%%%%%%%%%%%%

% \vspace{-3mm}
\section{Related Work}
\label{sec:related_work}

\textbf{LLM-based dialogue generation.} Recent advancements in LLMs, such as GPT \cite{achiam2023gpt}, Llama \cite{dubey2024llama}, DeepSeek \cite{guo2025deepseek} and Gemini \cite{comanici2025gemini} series, enable the generation of contextually relevant and coherent human-style dialogues from user-defined textual prompts. These LLMs can be categorised into three groups: text-only models \cite{guo2025deepseek,labs2025mercury} that interact with users via texts only, text-audio models \cite{fang2025llama,sharma2025indicsynth} supporting both textual and speech interactions, and multi-modal models that can understand and output text-audio \cite{dihan2025eyes, ghosh2024gama}, text-image \cite{glm2024chatglm,comanici2025gemini} and text-video \cite{liu2025ola,llavamini2025} modalities. However, to the best of our knowledge, none of the existing models can directly synthesise diverse audio-visual interactive and responsive clip pairs from a single user-defined prompt.

\textbf{Face video synthesis.} Existing face video generation approaches (Fig. \ref{fig:teaser}) can be roughly categorised into: (1) Condition-driven face video generation, (2) Face video editing, (3) Facial reaction generation, and (4) Interactive face video pair generation. Specifically, \textbf{condition-driven face video generation} methods aim to produce coherent facial animations from a face image \cite{xiang2025expressive} or 3D face data \cite{song2024talkingstyle} conditioned solely on texts \cite{guo2023animatediff,wang2025omnitalker}, audio \cite{jang2024faces,choi2024text} or other pre-defined attributes (e.g., action units \cite{luo2024reactface,he2025synfer} or driving motion fields \cite{guo2024sparse,zhao2025synergizing}). Alternatively, \textbf{face video editing} methods modify the given face videos through pre-defined or learned facial attributes, enabling expressive and realistic changes in identity \cite{bai2024bring} or emotions \cite{ma2024follow}. For example, talking face methods frequently utilise pre-defined speech \cite{zhang2023sadtalker,tan2024edtalk} or textual transcripts \cite{cui2025hallo4} to semantically modify face videos with synchronised lip behaviours \cite{li2024latentsync}, facial expressions \cite{tan2024edtalk} or enhanced hair-face consistency \cite{deng2025degstalk}. However, these methods can only output independent, non-interactive face videos defined by the given conditions.
% without the capability to generate interactive human-human interaction face videos that consider their conversational partner's behaviours.
Alternatively, an increasing number of methods \cite{song2023react2023,tran2024dim,liu2024one,song2025react,zhu2024perfrdiff,luo2025reactdiff,mao2025scattering,wang2025explaining,huang2025multiple,nguyen2024vector}  have been developed to generate diverse but appropriate interactive \textbf{facial reaction videos} conditioned on the input human audio-visual behaviours, although they still fail to synthesise paired interactive audio-visual clips. Besides, recent \textbf{interactive talking face generation} methods such as DialogueNeRF \cite{yan2024dialoguenerf}, DualTalk \cite{peng2025dualtalk}, Audio2Photoreal \cite{ng2024audio}, INFP \cite{zhu2025infp}, AV-Flow \cite{chatziagapi2025av} and ChatAnyone \cite{qi2025chatanyone}, enable direct generation of paired verbally interactive face videos. However, these interactive face videos are generated from pre-defined interactive verbal/language behaviours without semantically considering non-verbal facial and vocal behaviours in triggering/responding to their conversational partners' behaviours, and thus these methods fail to generate both verbally and non-verbally interactive outputs. Concurrent works TAVID \cite{kim2025tavid} and JAM-Flow \cite{kwon2025jam} explore related joint generation tasks but require pre-defined scripts or audio inputs. Compared with these methods, CHAT targets the setting of generating complete, mutually responsive DIAD clips from a \textbf{single anonymous scenario prompt} without any pre-defined conditions.

%%%%%%%%%%%%%%%%%%%%%%%%%%%%%%%%%%%%%%%%%%%%
%%%          Task Definition            %%%%
%%%%%%%%%%%%%%%%%%%%%%%%%%%%%%%%%%%%%%%%%%%%

\section{Task Definition}
\label{sec:task}

\noindent The DIADG task aims to develop a machine learning model $\mathcal{H}$ that can generate multiple different DIAD clips $D_1, D_2, \dots, D_N$ from a textual prompt $\mathcal{P}$. Here, $\mathcal{P}$ provides a de-identified summary of a dyadic interaction scenario where two subjects chat with each other. This DIADG task can be formally formulated as:
\begin{equation}
    D_1, D_2, \cdots, D_N = \mathcal{H}(\mathcal{P}),
\label{eq:task}
\end{equation}
where each DIAD clip $D_n = \{(V_n^{i}, A_n^{i}), (V_n^{j}, A_n^{j})\}, n \in \{1, \cdots, N\}$ contains realistic, temporally coherent, synchronised and mutually responsive speeches and facial behaviours expressed by two arbitrary subjects $S^i$ and $S^j$ engaged in a natural dyadic dialogue.

%%%%%%%%%%%%%%%%%%%%%%%%%%%%%%%%%%%%%%%%%%%%
%%%           Methodology               %%%%
%%%%%%%%%%%%%%%%%%%%%%%%%%%%%%%%%%%%%%%%%%%%

\section{Methodology}
\label{sec:method}

\subsection{CHAT Framework}
\label{subsec:framework}

\begin{figure}[t]
    \centering
    \includegraphics[width=\linewidth]{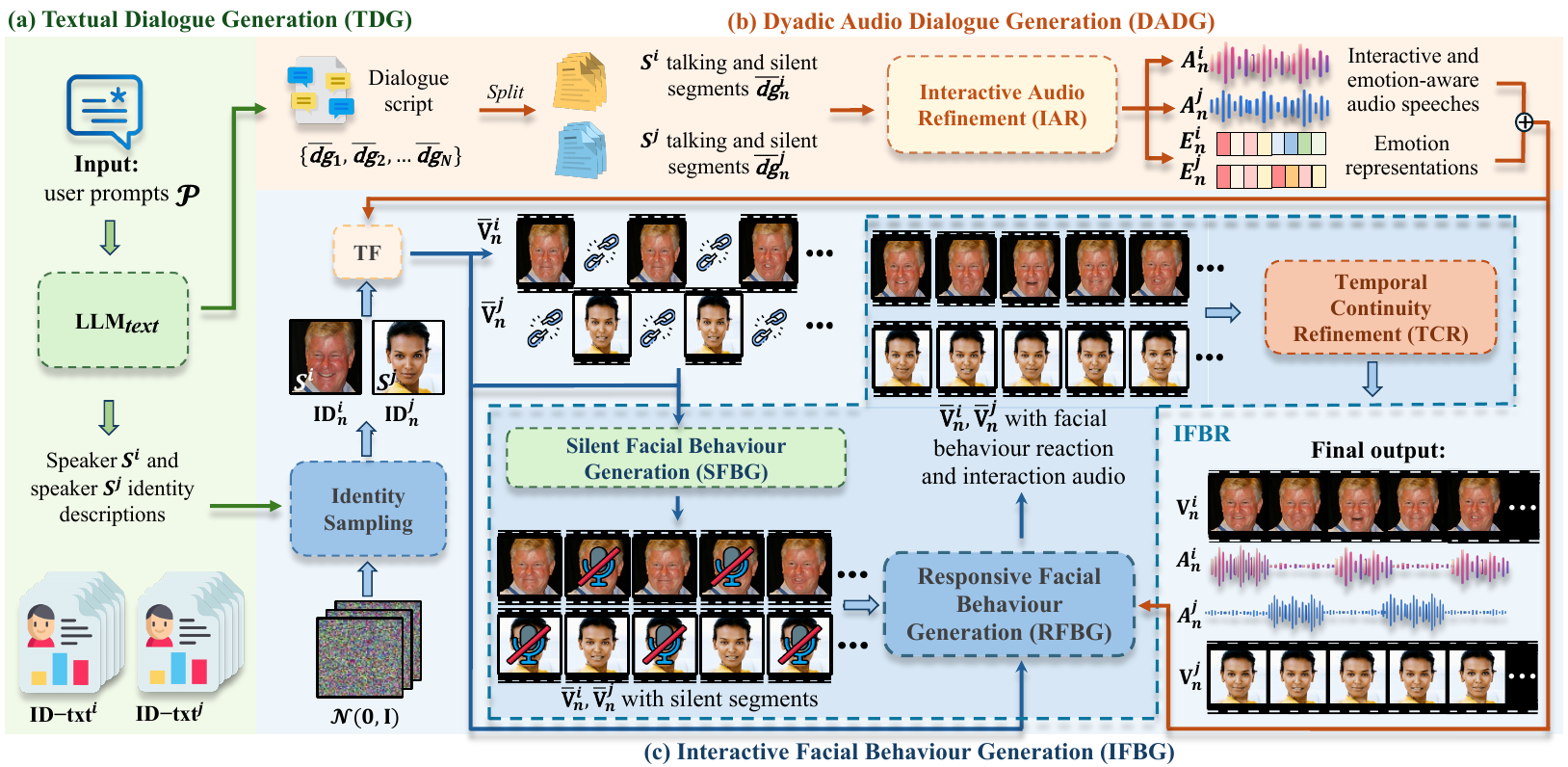}
    \caption{\textbf{The CHAT pipeline.} \textbf{(a)} TDG generates textual dialogue pairs and identity descriptors from a single user prompt $\mathcal{P}$; \textbf{(b)} DADG converts each dialogue into interactive, emotion-aware audio pairs via the IAR block (Sec. \ref{subsec:IAR}); \textbf{(c)} IFBG synthesises mutually responsive face video pairs using a talking face model and the IFBR block (Sec. \ref{subsec:IFBR}). All DIAD clips are assembled from DADG and IFBG outputs.}
    \label{fig:framework}
\end{figure}

\noindent Our \textbf{C}onversational \textbf{H}uman \textbf{A}udio-visual \textbf{T}alking (\textbf{CHAT}) Dialogue Generation is a DIADG solution that can generate diverse, contextually appropriate, temporally coherent, and mutually responsive dyadic interactive audio-visual dialogues (DIADs) $D_1, D_2, \cdots, D_N$ from a single user-defined textual prompt $\mathcal{P}$.
Throughout Sec. \ref{sec:method}, $\bar{x}$ denotes initially generated values, $\hat{x}$ denotes LLM-refined values, and $\tilde{x}$ denotes diffusion-refined values. As illustrated in Fig. \ref{fig:framework}, it comprises three core modules: Textual Dialogue Generation (\textbf{TDG}), Dyadic Audio Dialogue Generation (\textbf{DADG}), and Interactive Facial Behaviour Generation (\textbf{IFBG}). This modular design allows individual components (e.g., $\text{LLM}_\text{text}$) to be replaced or updated independently without retraining the full pipeline, while a two-stage scheme, which first pre-trains the individual modules and then jointly optimises the trainable blocks end-to-end with the LLMs and text-to-speech (TTS) models frozen, ensures stable optimisation. Separating dialogue, audio and video into dedicated modules lets each stage build on a model that is already strong in its own domain, and keeps the trainable part of the pipeline small, which matters given the cost of video diffusion.

\textbf{Textual Dialogue Generation (TDG).} Our CHAT starts with the TDG module responsible for generating multi-turn dyadic dialogues and identity descriptions from $\mathcal{P}$. Given a textual prompt $\mathcal{P}$ describing the scenario and style of the dialogue, a large language model (LLM) $\text{LLM}_\text{text}$ is employed to produce $N$ diverse, contextually relevant dialogue scripts $\bar{\textbf{DG}} = \{\bar{\text{dg}}_1, \bar{\text{dg}}_2, \cdots, \bar{\text{dg}}_N \}$, with each dialogue also paired with two generated textual descriptions $\text{ID-txt}^i$ and $\text{ID-txt}^j$ representing the identities of two individuals $S^i$ and $S^j$ engaged in the conversation (e.g., $S^i$ is a thoughtful gentleman and $S^j$ is a vibrant lady).

To ensure robustness and avoid the performance degradation of the LLM in generating long multi-turn conversational dialogues, our TDG generates each dyadic conversation via the LLM conditioned on the speakers' identities, dialogue texts and sentiment/emotion through a user prompt. Here, each conversation is constrained to 5--10 turns, a range where LLMs maintain optimal coherence without degradation \cite{li2024streamingdialogue, koudounas2025deepdialogue}. Here, each turn may contain multiple sentences expressed from one speaker.

\textbf{Dyadic Audio Dialogue Generation (DADG).} Given each textual dialogue $\bar{\text{dg}}_n = \{\bar{\text{dg}}_n(1), \cdots, \bar{\text{dg}}_n(K)\}$ containing $K$ sentences, the DADG module first organises them as two sets $\bar{\text{dg}}_n^{i}$ and $\bar{\text{dg}}_n^{j}$ based on their orders: odd-numbered sentences $\bar{\text{dg}}_n^{i} = \big\{ \bar{\text{dg}}_n(1), \text{Si}^i_n(2), \bar{\text{dg}}_n(3), \text{Si}^i_n(4), \dots, \bar{\text{dg}}_n(K-1), \text{Si}^i_n(K) \big\}$ expressed by $S^i$, even-numbered sentences $\bar{\text{dg}}_n^{j} = \big\{ \text{Si}^j_n(1), \bar{\text{dg}}_n(2), \text{Si}^j_n(3), \bar{\text{dg}}_n(4), \dots, \text{Si}^j_n(K-1), \bar{\text{dg}}_n(K) \big\}$ expressed by $S^j$. Here, $\text{Si}^i_n(k)$ denotes a silent segment with the same temporal duration as $\bar{\text{dg}}_n(k)$ (i.e., $S^i$ keeps silent when $S^j$ expresses the $k$-th sentence). While directly converting each textual sentence to an audio speech disregards the interactive nature of the human-human interaction, our DADG involves an \textbf{Interactive Audio Refinement (IAR)} block (please refer to Sec. \ref{subsec:IAR} for details) that encodes each organised dialogue $\bar{\text{dg}}_n^{i}, \bar{\text{dg}}_n^{j}$ as an interactive and emotion-aware audio dialogue $\mathcal{A}_n = \{A_n^{i}, A_n^{j}\}$ conditioned on a user-defined prompt $\mathcal{P}_\text{ref}$ as: 
\begin{equation}
    A_n^{i}, A_n^{j}, E_n^i, E_n^j = \text{IAR}(\bar{\text{dg}}_n^{i}, \bar{\text{dg}}_n^{j}, \mathcal{P}_\text{ref}),
\end{equation}
where each speech sentence in $A_n^{i}$/$A_n^{j}$ is expressed with a specific emotion (defined by $E_n^i$, $E_n^j$) suitable for the context. In natural human-human conversation, although turn-taking is the fundamental structure and simultaneous speaking is a statistically rare edge case in most interactive scenarios, our IAR still occasionally inserts short-term interactive speech responses (e.g., ``Hmm'' and ``OK'') to silent segments to express the silent individual's verbal emotional responses when the conversational partner is speaking.

\textbf{Interactive Facial Behaviour Generation (IFBG).} Finally, the IFBG module generates a pair of mutually responsive, speech-aligned face videos ($V_n^i$ and $V_n^j$) from $A_n^i$ and $A_n^j$. It first extends a talking face model $\text{TF}$ \cite{aneja2024facetalk} to generate the initial talking facial behaviour segments (i.e., $\bar{V}_n^i = \{\bar{v}_n^i(1), \bar{v}_n^i(3), \cdots, \bar{v}_n^i(K-1) \}$ and $\bar{V}_n^j = \{\bar{v}_n^j(2), \bar{v}_n^j(4), \cdots, \bar{v}_n^j(K) \}$) corresponding to the given speech sentences, where the speech features are additionally concatenated with their corresponding emotion representations obtained by IAR to form a joint emotion-aware speech condition, which is injected into the talking face model to guide both accurate lip synchronisation and emotionally expressive facial animations. Here, an identity generation/sampling block \cite{papantoniou2024arc2face} converts the textual identity descriptors into sampled synthetic face images $\text{ID}_n^i$ and $\text{ID}_n^j$ from the texts $\text{ID-txt}^i$ and $\text{ID-txt}^j$ for defining the facial identities of $S^i$ and $S^j$, i.e., supporting diverse photorealistic and stylised identities. Based on both face identities, we propose an \textbf{Interactive Facial Behaviour Refinement (IFBR)} block to: (i) synthesise facial behaviours of silent segments; (ii) ensure all talking and silent facial behaviours are responsive to their conversational partner's audio-visual behaviours; and (iii) make the final refined facial behaviours $V_n^i$ and $V_n^j$ temporally coherent at the segment boundaries as: 
\vspace{-2mm}
\begin{equation}
      V_n^i, V_n^j = \text{IFBR}(\bar{V}_n^i, \bar{V}_n^j, A_n^i, A_n^j).
\end{equation}
Please refer to Sec. \ref{subsec:IFBR} for more details.

\textbf{CHAT-AVD-50k dataset.} CHAT-AVD-50k, a large-scale synthetic dataset for DIADG research, contains 50,000 diverse dyadic textual-audio-visual dialogue pairs (100,000 clips, 1388.9 hours) across 100,000 synthesised facial identities, with turn-level emotion, scenario, and speaker identity metadata. It covers casual, professional, emotional, academic, and social interaction scenarios.

\subsection{Interactive Audio Refinement}
\label{subsec:IAR}

\noindent The \textbf{IAR block} of the DADG module first applies an LLM ($\text{LLM}_\text{audio}$) to refine the obtained textual dialogue $\bar{\text{dg}}_n^{i}$ and $\bar{\text{dg}}_n^{j}$ using an audio refine prompt  $\mathcal{P}_\text{ref}$. This not only refines the contents of $\bar{\text{dg}}_n^{i}$ and $\bar{\text{dg}}_n^{j}$ as $\hat{\text{dg}}_n^{i}$ and $\hat{\text{dg}}_n^{j}$ by making extensive use of the second-person pronouns and conversational sentences, but also outputs:

\noindent \textbf{(i)} \textbf{a set of interactive words} (e.g., Hmm and OK) $\text{IW}_n = \{\text{iw}_n(1), \cdots, \text{iw}_n(K) \}$ allowing each individual to express short but contextually appropriate responses when the conversational partner speaks, making the dialogue more natural and interactive as: 
\begin{equation}
    \text{iw}_n(k) = \begin{cases}
        \text{random}(\mathcal{W}_\text{inter}) & \text{random}() < P_\text{inter} \\
        \text{silence} & \text{otherwise}
    \end{cases},
\end{equation}
where $\mathcal{W}_\text{inter} = \{\text{``uh-huh''}, \text{``hmm''}, \text{``okay''}, \text{``yeah''}\}$ and $P_\text{inter}$ denotes a pre-defined probability.

\noindent \textbf{(ii)} \textbf{a pair of emotion representations} $E_n^i = \{ {e_n^i}(1), \cdots, {e_n^i}(K) \}$ and $E_n^j = \{ {e_n^j}(1), \cdots, {e_n^j}(K) \}$, where each $e_n^i(k)$ is a structured descriptor from $\text{LLM}_\text{audio}$ encoding a discrete emotion category and continuous prosody attributes (rate, pitch, energy, pauses), embedded by the emotion encoder described below;

\noindent \textbf{(iii)} \textbf{time stamps} $\mathcal{T}_n^i$ and $\mathcal{T}_n^j$ for labelling the start and end time stamps of all sentences and interactive words expressed by $S^i$ and $S^j$;

\noindent \textbf{(iv)} \textbf{the sound environment} $\text{SE}_n$ (e.g., studio or open space) of the dialogue. Then, we insert all interactive words into their corresponding silent segments in $\hat{\text{dg}}_n^{i}$ and $\hat{\text{dg}}_n^{j}$, resulting in the final interactive dialogues $\text{dg}_n^{i}$ and $\text{dg}_n^{j}$ (written without a hat to denote the interactive-word-inserted sequences):
\begin{equation}
\begin{split}
    \text{dg}_n^{i} = \big\{\text{dg}_n(1), \text{iw}_n(2), \text{dg}_n(3),  \cdots,  \text{iw}_n(K) \big\}, \\
    \text{dg}_n^{j} = \big\{\text{iw}_n(1), \text{dg}_n(2), \text{iw}_n(3),  \cdots, \text{dg}_n(K) \big\}.
\end{split}
\end{equation}
As a result, the final obtained $\text{dg}_n^{i}$, $\text{dg}_n^{j}$, $E_n^i, E_n^j$, $\mathcal{T}_n^i$, $\mathcal{T}_n^j$ and $\text{SE}_n$ jointly represent an interactive human speech dialogue.

To generate interactive and emotion-aware audio speeches from the refined textual dialogues, each sentence/interactive word $\text{dg}_n^i(k) /  \text{iw}_n(k)$ and its corresponding emotion representation ${e_n^i}(k)$ are first encoded by a content Transformer encoder (TFM-C) and an emotion Transformer encoder (TFM-E) into a pair of latent representations of the same dimension. An element-wise sum then processes these representations to generate an emotion-aware dialogue representation $\mathbf{z}_n^i(k)$ as:
\begin{equation}
    \mathbf{z}_n^i(k) = \text{Sum}(\text{TFM-C}(\text{dg}_n^i(k)), \text{TFM-E}({e_n^i}(k))).
\end{equation}
Then, the obtained $\{\mathbf{z}_n^i(k)\}_{k=1}^K$ corresponding to all sentences/interactive words expressed by $S^i$ 
%%% IAR details
are used, together with the time stamps $\mathcal{T}_n^i$ and the sound environment $\text{SE}_n$, to condition an emotional speech synthesiser that produces the interactive, emotion-aware audio waveform segments ${A_n^i(k)}_{k=1}^K$.

\subsection{Interactive Facial Behaviour Refinement}
\label{subsec:IFBR}

\noindent The \textbf{IFBR block} refines the independently generated \emph{non-interactive} talking facial behaviour segments $\bar{V}_n^i$ and $\bar{V}_n^j$ as a pair of temporally coherent and mutually responsive face videos $V_n^i$ and $V_n^j$ via the following blocks.

\textbf{Silent Facial Behaviour Generation (SFBG).} This block generates each initial silent facial behaviour segment $\bar{v}_n^i(k)$/$\bar{v}_n^j(k+1)$  when $S^i$/$S^j$ are not speaking, reflecting temporally and semantically coherent facial behaviours in the context of its preceding and succeeding talking facial behaviour segments $\bar{v}_n^i(k-1)$ and $\bar{v}_n^i(k+1)$.
% their active listening, emotional processing, and anticipatory responses to the conversational partner's verbal and non-verbal behaviours expressed at the same period.
Specifically, the initial silent segment $\bar{v}_n^i(k)$ containing realistic facial behaviours (e.g., nods and blinks) is generated conditioned on $S^i$'s facial identity image $\text{ID}_n^i$ as well as $\bar{v}_n^i(k-1)$ and $\bar{v}_n^i(k+1)$ via a video diffusion model $\text{SilentDiff}$ as:
\begin{equation}
    \bar{v}_n^i(k) = \text{SilentDiff}(\bar{v}_n^i(k-1), \bar{v}_n^i(k+1), \text{ID}_n^i).
\end{equation}
SilentDiff inherits the same architecture from \cite{cui2025hallo3} but is end-to-end optimised with other modules/blocks within our framework. This also maintains rough temporal continuity and semantic coherence between $\bar{v}_n^i(k)$ and its adjacent talking facial segments $\bar{v}_n^i(k-1)$/$\bar{v}_n^i(k+1)$.

\begin{figure}[t]
    \centering    
    \includegraphics[width=\linewidth]{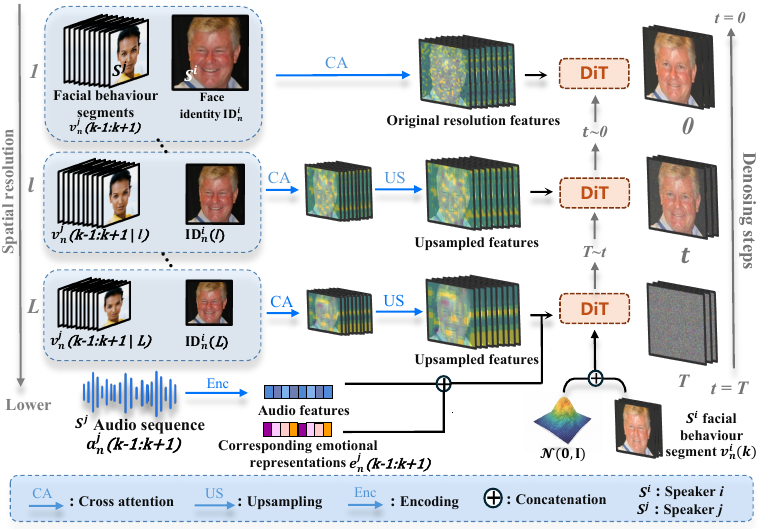}
    \caption{Illustration of the proposed RFBG sub-block.}
    \label{fig:resolutions}
\end{figure}

\textbf{Responsive Facial Behaviour Generation (RFBG).} This block employs a video diffusion transformer to refine the initial silent and talking facial behaviour segments $\bar{v}_n^i(k)$/$\bar{v}_n^i(k{-}1)$ as facial behaviour segments $\tilde{v}_n^i(k)$/$\tilde{v}_n^i(k{-}1)$ that are responsive to the conversational partner's audio-visual behaviours expressed at the same period. Given $S^i$'s facial behaviour $\bar{v}_n^i(k)$, the conditions used in RFBG include $S^j$'s preceding, current and succeeding audio-visual behaviours $\bar{v}_n^j(k{-}1:k{+}1)$ and $a_n^j(k{-}1:k{+}1)$ expressed from turn $k-1$ to $k+1$, which are injected into the diffusion transformer at multiple denoising steps. As illustrated in Fig. \ref{fig:resolutions}, the face identity $\text{ID}_n^i$ and $S^j$'s facial behaviours $\bar{v}_n^j(k-1:k+1)$ are spatially downsampled to $L$ scales as ${\text{ID}_n^i(l)}_{l=1}^L$ and $\bar{v}_n^j(k{-}1:k{+}1 | l)_{l=1}^L$ (i.e., higher $l$ corresponds to lower spatial resolution), which are then fused via learnable cross-attention (CA) blocks at $L$ scales as:
\begin{equation}
    \mathcal{C}_n^j(k{-}1{:}k{+}1 | l) = \text{CA}({\bar{v}_n^j(k{-}1:k{+}1 | l), \text{ID}_n^i(l)}).
\end{equation}
This feature extraction enables the capture of both multi-scale coarse-level (e.g., low-resolution conditions generally capture shape and global facial motions) and fine-grained (e.g., high-resolution conditions typically capture facial details such as micro-expressions and local muscle movements) facial behaviours expressed by $S^j$, while forcing the diffusion transformer to generate facial behaviours of identity $\text{ID}_n^i$. Then, the obtained $L$ conditions ${\mathcal{C}_n^j(k{-}1{:}k{+}1 | l)}_{l=1}^L$ (i.e., each is a 2D feature map sequence) are upsampled to the same spatial resolution as the original $\bar{v}_n^j(k{-}1:k{+}1)$/$\text{ID}_n^i$. We then gradually integrate conditions obtained from lower spatial resolutions into early denoising steps and conditions of higher spatial resolutions into later denoising steps, as the early steps of the diffusion process typically model overall facial behaviour outlines, while later steps refine facial details \cite{hertz2022prompt,wang2023diffusion} as:
\begin{equation}  
    \tilde{v}_n^i(k) = \text{Diff}\big(\bar{v}_n^i(k) \mid \Phi_k^{i,j}(t), a_n^j(k{-}1:k{+}1), e_n^j(k{-}1:k{+}1) \big),
\end{equation}
where $\Phi_k^{i,j}(t) = \bigcup_{l=l(t)}^{L} \mathcal{C}_n^j(k{-}1{:}k{+}1 | l)$. Particularly, $l(t) = \max(1,\, \lceil L\,t/T \rceil)$ denotes that at denoising step $t$, the facial behaviour conditions $\mathcal{C}_n^j(k{-}1{:}k{+}1 | l)$ from scale $l(t)$ to $L$ are injected into the diffusion process. This ensures that the conditions capturing $S^j$'s global facial motions guide early denoising steps to generate responsive facial behaviour outlines of $\tilde{v}_n^i(k)$, whilst the fine-grained conditions are progressively integrated into later denoising steps to help complete details of $\tilde{v}_n^i(k)$. At the first denoising step, we also inject the audio representation learned from $a_n^j(k{-}1:k{+}1)$ and its emotion representation $e_n^j(k{-}1:k{+}1)$ obtained by IAR via cross-attention as a condition.

\textbf{Temporal Continuity Refinement (TCR).} Although the SFBG block maintains rough temporal continuity between adjacent facial segments, segment-level processing conducted by the RFBG block may still result in visual discontinuities at facial segment boundaries. To address this, we utilise a Gaussian-weighted boundary blending strategy \cite{niklaus2020softmax} to ensure: (1) smooth and perceptually imperceptible facial behaviour transition between facial segment boundaries; and (2) impact on only a narrow temporal window within $2W$ frames centred at each boundary (e.g., empirically set to only $W=10$ frames, whereas each facial behaviour segment contains 100 frames), as the Gaussian weighting function rapidly decays towards zero outside the boundary region. Importantly, this block conducts \textbf{an identical transform to both $S^i$ and $S^j$ facial behaviours} at the same time window to preserve their established mutual responsiveness. Given a pair of adjacent facial behaviour segments $\tilde{v}_n^i(k-1)$ and $\tilde{v}_n^i(k)$, this block applies temporal blending to the last $W$ frames of $\tilde{v}_n^i(k-1)$ and the first $W$ frames of $\tilde{v}_n^i(k)$ as:
\begin{equation}
\begin{aligned}
v_n^i(k-1,\, F_{k-1}-\tau+1) &=
[1-\alpha(\tau)]\, \tilde{v}_n^i(k-1,\, F_{k-1}-\tau+1) \\
& \quad + \alpha(\tau)\, \tilde{v}_n^i(k,\, \tau), \\[4pt]
v_n^i(k,\, \tau) &= 
\alpha(\tau)\, \tilde{v}_n^i(k-1,\, F_{k-1}-\tau+1) \\
& \quad + [1-\alpha(\tau)]\, \tilde{v}_n^i(k,\, \tau),
\end{aligned}
\end{equation}
where $\quad 1 \le \tau \le W$ and $F_{k-1}$ is the total number of frames in segment $k-1$, and $\alpha(\tau) = \tfrac{1}{2}\exp(-(\tau-1)^2/2\sigma^2)$ is a half-Gaussian blending weight, equal to $0.5$ at the boundary and decaying outwards, with $\sigma = W/3$ following the 3-sigma rule. Since the transforms on both segments share the same $\alpha(\tau)$, this symmetric design preserves the established mutual responsiveness between $S^i$ and $S^j$ facial behaviours.

%%%%%%%%%%%%%%%%%%%%%%%%%%%%%%%%%%%%%%%%%%%%
%%%              Experiments            %%%%
%%%%%%%%%%%%%%%%%%%%%%%%%%%%%%%%%%%%%%%%%%%%

\section{Experiments}
\label{subsec:experiment}

%%% 
\subsection{Experimental Settings}

\noindent \textbf{Implementation details.} The LLMs used in TDG and DADG modules are the Gemini models \cite{comanici2025gemini}, while the emotional audio generation model in IAR is a pre-trained TTS \cite{casanova2024xtts}. 
%with temperature 0.7, top-p 0.9, and repetition penalty 1.1, pre-trained on over 100,000 hours of real English speech data using AdamW optimiser.
%the REACT 2024 \cite{song2024react} dataset using AdamW optimiser with learning rate 1e-4, weight decay 1e-6, and batch size 32 for 100,000 training steps.
We pre-train SFBG and RFBG blocks on the HDTF \cite{zhang2021flow} dataset and REACT 2024 dataset \cite{song2024react} using AdamW optimiser respectively, where the diffusion model trained for RFBG is set to have 50 denoising steps.
%learning rate 1e-5, weight decay 1e-2, and batch size 4 for maximum 30,000 training steps. 
% The RFBG block is trained on the REACT 2024 dataset \cite{song2024react} with AdamW optimiser, 
%with learning rate 1e-4, weight decay 1e-4, and batch size 4 for maximum 500 epochs. 
% where the diffusion process uses 1000 training timesteps with a cosine noise schedule and 50 inference steps.
Then, the talking face model $\text{TF}$ in IFBG as well as SFBG and RFBG blocks are end-to-end trained, while other blocks (LLMs and TTS models) are frozen. We compare our CHAT with competitors by utilising them to generate 1000 audio-visual dialogues, with each containing 5--10 conversational turns covering diverse topics (e.g., casual conversations and professional discussions).

\begin{figure}[t]
    \centering
    \includegraphics[width=\linewidth]{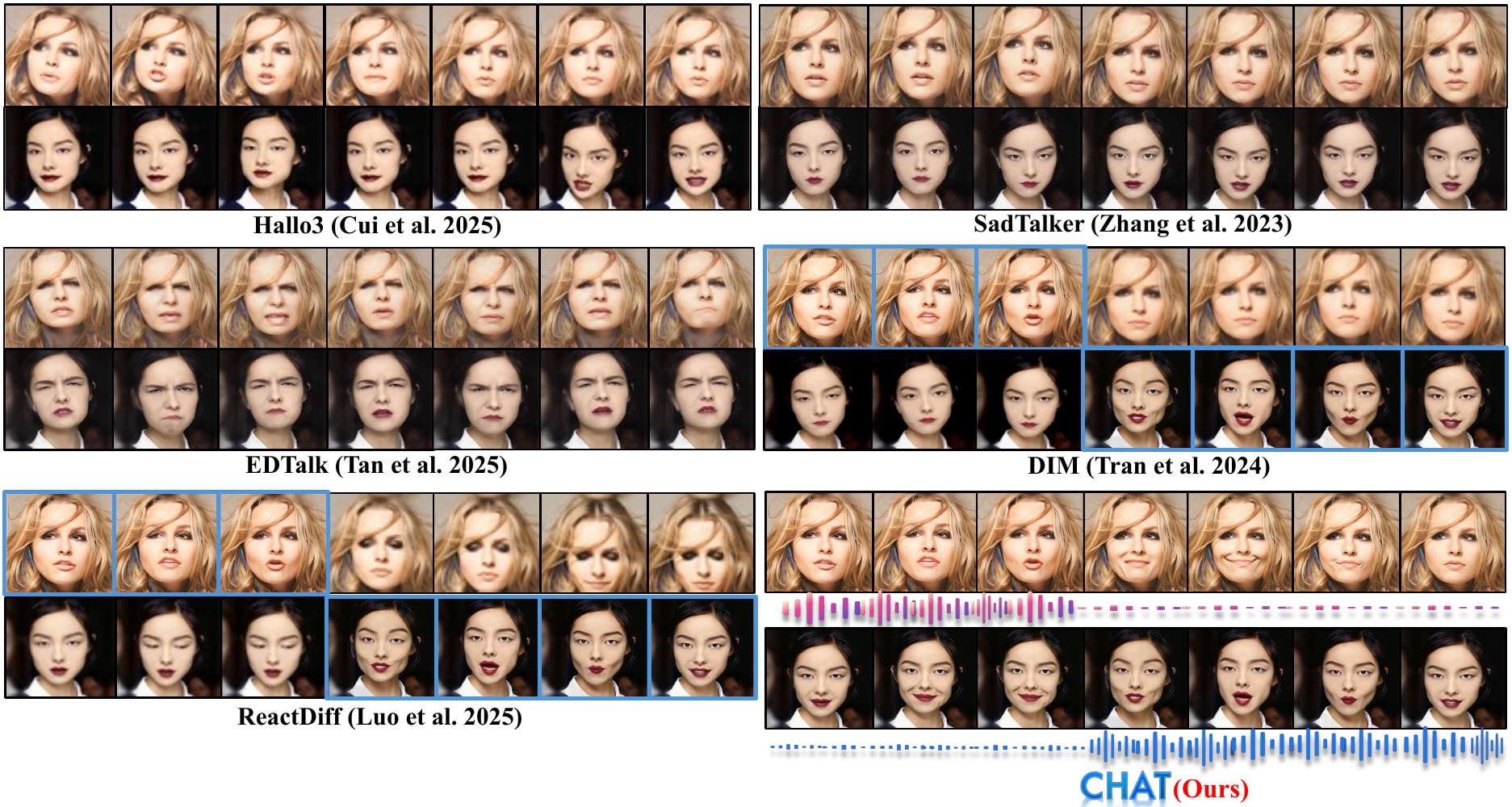}
    \caption{\textbf{Qualitative comparison}. Talking-face methods (Hallo3, EDTalk, SadTalker) used twice yield non-responsive pairs; FRG methods (DIM, ReactDiff) take user-provided speaker video (blue frames). CHAT generates mutually responsive audio-visual dialogue pairs.}
    \label{fig:qualitative}
\end{figure}

\noindent \textbf{Evaluation metrics.} We assess four aspects: \textbf{Visual quality} via FID \cite{heusel2017gans} and FVD \cite{unterthiner2018towards}; \textbf{AV synchronisation} via LSE-C \cite{prajwal2020lip} and LSE-D \cite{li2024latentsync}; \textbf{Interactive responsiveness} via FRCorr and FRDiv \cite{song2023react2023} (appropriateness and diversity of generated behaviours); \textbf{Identity preservation} via CSIM (ArcFace \cite{deng2019arcface}) and LPIPS \cite{zhang2018unreasonable}. Ablation studies additionally report MCD \cite{kubichek1993mel}, Emo-Acc \cite{ravanelli2021speechbrain}, and ViSQOL \cite{chinen2020visqol}. A 60-participant user study provides subjective quality assessment on a 5-point Likert scale.

\setlength{\abovecaptionskip}{4pt}
\setlength{\belowcaptionskip}{1pt}
\setlength{\floatsep}{5pt plus 1pt minus 1pt}
\setlength{\textfloatsep}{7pt plus 1pt minus 1pt}
\begin{table}[t]
\centering
\caption{\textbf{Quantitative results.} Comparison of CHAT and baselines on evaluation set (mean of 1000 samples). The best result is in \textbf{bold} and the second best in \underline{underline}.}
\resizebox{\linewidth}{!}{
\small
\begin{tabular}{ccccccccc}
\toprule
\multirow{2}{*}{\textbf{Method}} 
& \multicolumn{2}{c}{\textbf{Visual Quality}} 
& \multicolumn{2}{c}{\textbf{AV Sync}} 
& \multicolumn{2}{c}{\textbf{Reaction ($\times10^{-2}$)}} 
& \multicolumn{2}{c}{\textbf{ID Preservation}} \\
\cmidrule(lr){2-3} \cmidrule(lr){4-5} \cmidrule(lr){6-7} \cmidrule(lr){8-9}
& FID $\downarrow$ & FVD $\downarrow$ 
& LSE-C $\uparrow$ & LSE-D $\downarrow$ 
& FRCorr $\uparrow$ 
& FRDiv $\uparrow$ 
& CSIM $\uparrow$ & LPIPS $\downarrow$ \\
\midrule
Hallo3 \cite{cui2025hallo3} & 20.56 & \textbf{362.23} & 4.61 & 10.41 & N/A & N/A & 0.79 & 0.55 \\
SadTalker \cite{zhang2023sadtalker} & 22.53 & 385.51 & \underline{6.85} & \underline{8.17} & N/A & N/A & 0.81 & 0.48 \\
DIM \cite{tran2024dim} & 36.52 & 460.35 & 6.82 & 8.89 & 31.02 & 12.55 & \underline{0.83} & 0.52 \\
EDTalk \cite{tan2024edtalk} & \underline{18.74} & 619.92 & 5.62 & 9.61 & N/A & N/A & 0.78 & \textbf{0.36} \\
ReactDiff \cite{luo2025reactdiff} & 21.36 & 386.22 & 5.23 & 9.02 & \underline{48.05} & \underline{15.31} & 0.77 & 0.58 \\
\midrule
\rowcolor{gray!10}
\textbf{CHAT (\textcolor{red}{Ours})} & \textbf{17.33} & \underline{365.03} & \textbf{6.89} & \textbf{8.07} &  \textbf{50.02} & \textbf{15.34} & \textbf{0.85} & \underline{0.38} \\
\bottomrule
\end{tabular}
}
\label{tab:quantitative_results}
\end{table}

\begin{table}[t]
\centering
\caption{\textbf{Qualitative user study results.} Mean opinion scores for each method and evaluation dimension. The best result is in \textbf{bold} and the second best in \underline{underline}.}
% \resizebox{\linewidth}{!}{
\small
\setlength{\tabcolsep}{5pt}
\begin{tabular}{l|ccccc}
\hline
\textbf{Method} & \textbf{Visual} & \textbf{Audio} & \textbf{Sync} & \textbf{Expr.} & \textbf{Interact.} \\
\hline
Hallo3 \cite{cui2025hallo3}          & \underline{4.5} & 3.5 & \underline{3.9} & 3.8 & 2.5 \\
SadTalker \cite{zhang2023sadtalker}       & 2.5 & \underline{3.7} & 1.2 & 2.0 & 1.3  \\
DIM \cite{tran2024dim}            & 3.7 & 3.1 & 3.4 & 2.8 & \underline{3.6} \\
EDTalk \cite{tan2024edtalk}         & 2.2 & 3.6 & 2.3 & 2.8 & 2.7 \\
ReactDiff \cite{luo2025reactdiff}      & 4.2 & 3.5 & 3.8 & \underline{4.1} & 3.2 \\
\hline
\rowcolor{gray!10}
\textbf{CHAT (\textcolor{red}{Ours})} & \textbf{4.6} & \textbf{4.8} & \textbf{4.6} & \textbf{4.6} & \textbf{4.8} \\
\hline
\end{tabular}
% }
\label{tab:user_study}
\end{table}

\begin{table*}[t]
\centering
\small
\caption{\textbf{Ablation quantitative analysis results.} Performance comparison of CHAT variants (mean of 1000 samples). All variants include three base modules: TDG, DADG, and IFBG. \cmark{} indicates enabled, and \xmark{} indicates disabled. (i) without entire IAR (but keep time stamps); (ii) without interactive words; (iii) without emotion representations; (iv) without sound environment; (v) without entire IFBR; (vi) without SFBG (replaced by interpolation); (vii) without RFBG; and (viii) without TCR. A dash marks an audio metric that is not applicable, as the ablated module leaves the audio path unchanged.}
\resizebox{\linewidth}{!}{
\begin{tabular}{l|ccc|ccc||ccc|cccc|cc|cc}
\toprule
\multirow{2.5}{*}{\textbf{Variant}} 
& \multicolumn{3}{c|}{\textbf{IAR Components}} 
& \multicolumn{3}{c||}{\textbf{IFBR Components}}
& \multicolumn{3}{c|}{\textbf{Audio Quality}}
& \multicolumn{2}{c}{\textbf{Visual Quality}} 
& \multicolumn{2}{c}{\textbf{AV Sync}} 
& \multicolumn{2}{c}{\textbf{Reaction ($\times10^{-2}$)}} 
& \multicolumn{2}{c}{\textbf{ID Preservation}} \\
\cmidrule(lr){2-4} \cmidrule(lr){5-7} \cmidrule(lr){8-10} \cmidrule(lr){11-12} \cmidrule(lr){13-14} \cmidrule(lr){15-16} \cmidrule(lr){17-18}
& IW & Emo & SE
& SFBG & RFBG & TCR
& MCD$\downarrow$ & Emo-Acc$\uparrow$ & ViSQOL$\uparrow$
& FID$\downarrow$ & FVD$\downarrow$ 
& LSE-C$\uparrow$ & LSE-D$\downarrow$ 
& FRCorr$\uparrow$ & FRDiv$\uparrow$
& CSIM$\uparrow$ & LPIPS$\downarrow$ \\
\midrule
(i) w/o IAR     & \xmark & \xmark & \xmark & \cmark & \cmark & \cmark & 6.23 & 41.2\% & 2.97 & 21.34 & 456.78 & 6.03 & 9.45 & 32.45 & 7.89 & 0.74 & 0.53 \\
(ii) w/o IW            & \xmark & \cmark & \cmark & \cmark & \cmark & \cmark & 5.12 & 67.8\% & \underline{3.89} & 18.67 & 398.23 & 6.77 & 8.67 & 47.12 & 9.34 & 0.81 & 0.41 \\
(iii) w/o emotion       & \cmark & \xmark & \cmark & \cmark & \cmark & \cmark & 9.12 & 38.3\% & 2.67 & 23.45 & 423.56 & 6.12 & 9.23 & 28.34 & 5.67 & 0.69 & 0.59 \\
(iv) w/o SE            & \cmark & \cmark & \xmark & \cmark & \cmark & \cmark & \underline{4.78} & \underline{76.2\%} & 3.78 & \underline{18.45} & 385.67 & 6.72 & \underline{8.34} & 48.23 & 14.12 & \underline{0.84} & \underline{0.40} \\
\midrule
(v) w/o IFBR    & \cmark & \cmark & \cmark & \xmark & \xmark & \xmark & -- & -- & -- & 38.67 & 623.45 & 5.72 & 10.23 & 28.56 & 6.78 & 0.62 & 0.71 \\
(vi) w/o SFBG          & \cmark & \cmark & \cmark & \xmark & \cmark & \cmark & -- & -- & -- & 20.23 & \underline{375.89} & \underline{6.78} & 8.94 & \underline{49.45} & 10.23 & 0.80 & 0.44 \\
(vii) w/o RFBG          & \cmark & \cmark & \cmark & \cmark & \xmark & \cmark & -- & -- & -- & 20.34 & 412.56 & 6.45 & 8.88 & 25.78 & 4.56 & 0.76 & 0.42 \\
(viii) w/o TCR           & \cmark & \cmark & \cmark & \cmark & \cmark & \xmark & -- & -- & -- & 18.56 & 421.34 & 6.67 & 9.12 & 48.67 & \underline{14.23} & 0.82 & 0.43 \\
\midrule
\rowcolor{gray!10}
\textbf{Full CHAT} & \cmark & \cmark & \cmark & \cmark & \cmark & \cmark & \textbf{4.23} & \textbf{78.4\%} & \textbf{3.98} & \textbf{17.33} & \textbf{365.03} & \textbf{6.89} & \textbf{8.07} & \textbf{50.02} & \textbf{15.34} & \textbf{0.85} & \textbf{0.38} \\
\bottomrule
\end{tabular}
}
\label{tab:ablation_quantitative}
\end{table*}

\subsection{Quantitative Results}
\noindent Since no existing method directly addresses DIADG, we compare CHAT with recent methods for related tasks: \textbf{Hallo3} \cite{cui2025hallo3} and \textbf{SadTalker} \cite{zhang2023sadtalker} for audio-driven talking face generation; \textbf{DIM} \cite{tran2024dim} and \textbf{ReactDiff} \cite{luo2025reactdiff} for facial reaction generation; and \textbf{EDTalk} \cite{tan2024edtalk} for emotional talking head synthesis. Audio-visual dialogue generation methods \cite{chatziagapi2025av,zhu2025infp,yan2024dialoguenerf,siniukov2025DiTaiListener,qi2025chatanyone,kwon2025jam,kim2025tavid} are excluded because code is unavailable. For fair comparison, all methods receive the same prompts/audio; since competitors are limited to 12-second clips while CHAT can generate substantially longer conversations, we standardise evaluation to the first ten seconds. Audio-driven baselines (Hallo3, SadTalker, EDTalk) receive CHAT's generated audio $A_n$ and identity $\text{ID}_n$, while FRG baselines (DIM, ReactDiff) receive CHAT's generated speaker video. Table \ref{tab:quantitative_results} shows that CHAT achieves the best audio-visual synchronisation, interactiveness, FID and CSIM, while ranking second in FVD and LPIPS. Although these baselines were developed for related rather than identical tasks, the results still indicate that CHAT can produce synchronised, identity-consistent, and interactive audio-visual clip pairs, providing a competitive reference for this newly introduced task. FRCorr and FRDiv measure facial reaction appropriateness and diversity, and do not fully reflect turn-taking or semantic coherence at the dialogue level.

\subsection{Qualitative Results}

\noindent We conducted a user study with 60 participants (aged 20--55, from academia and industry across the USA, the UK, Singapore and China), who evaluated randomly ordered video pairs on five aspects (visual realism, audio quality, AV sync, expressiveness, interactivity; 1--5 Likert scale) for both the main comparison (Table \ref{tab:user_study}) and eight ablation variants (Table \ref{tab:ablation_user_study}). Participants rated CHAT highest on all five aspects. Fig. \ref{fig:qualitative} also visually compares a dialogue generated by our CHAT and competitors. Whilst competitors exhibit visual artefacts around the mouth and generate strained expressions, our CHAT produces natural smiling facial reactions with higher fidelity and expressiveness. This is because CHAT is specifically designed for the DIADG task, while competitors were only designed for similar but different tasks. CHAT generates the two clips as a responsive pair, so the listening face reacts to the speaker within the same clip. The talking-face baselines animate each face in isolation, and the reaction baselines produce facial motion without the accompanying speech. These differences are consistent with the higher interactivity and synchronisation scores of CHAT in Table \ref{tab:quantitative_results} and Table \ref{tab:user_study}.

\subsection{Ablation Studies}
\label{subsec:abl}

\noindent We conducted a series of ablation studies (Table \ref{tab:ablation_quantitative}) to evaluate our CHAT.

\noindent \textbf{IAR ablation.} Removing IAR degrades audio quality consistently; eliminating emotion-aware processing causes the largest drop. IAR also underpins the mutual responsiveness of the generated facial reactions.

\noindent \textbf{IFBR ablation.} Removing IFBR eliminates both listener silence behaviours (SFBG) and partner-responsive refinement (RFBG), causing severe degradation: FID increases by 123.1\% and FVD by 70.8\%, confirming that facial behaviour refinement is fundamental. Removing RFBG alone leads to large interactivity loss, demonstrating its necessity for mutually responsive synthesis. Because these variants leave the DADG audio path unchanged, we omit their audio metrics and report only the visual, synchronisation, responsiveness, and identity results.

\noindent \textbf{User study on ablation variants.} Table \ref{tab:ablation_user_study} shows the same pattern. Removing IAR lowers the audio quality score by 2.0 and interactivity by 2.3, whereas removing IFBR lowers visual realism by 2.2 and expressiveness by 2.1. The two blocks therefore play complementary roles.

\begin{table}[t]
\centering
\caption{\textbf{Ablation user study results.} Mean opinion scores (1--5); base modules (TDG, DADG, IFBG) always included. Best in \textbf{bold}, second best \underline{underlined}.}
\resizebox{\linewidth}{!}{
\small
\begin{tabular}{l|ccc|ccc||ccccc}
\hline
\multirow{2}{*}{\textbf{Variant}} 
& \multicolumn{3}{c|}{\textbf{IAR Components}} 
& \multicolumn{3}{c||}{\textbf{IFBR Components}}
& \multicolumn{5}{c}{\textbf{User Study Scores (1--5)}} \\
\cmidrule(lr){2-4} \cmidrule(lr){5-7} \cmidrule(lr){8-12}
& IW & Emo & SE
& SFBG & RFBG & TCR
& Vis. & Aud. & Sync & Expr. & Inter. \\
\hline
w/o IAR (all)     & \xmark & \xmark & \xmark & \cmark & \cmark & \cmark & 3.2 & 2.8 & 3.1 & 2.3 & 2.5 \\
w/o IW            & \xmark & \cmark & \cmark & \cmark & \cmark & \cmark & 3.6 & 3.2 & 3.4 & 3.5 & 3.2 \\
w/o Emotion       & \cmark & \xmark & \cmark & \cmark & \cmark & \cmark & 2.9 & 3.1 & 2.8 & 1.7 & 2.8 \\
w/o SE            & \cmark & \cmark & \xmark & \cmark & \cmark & \cmark & 3.6 & 3.6 & 3.5 & 3.6 & 4.1 \\
\hline
w/o IFBR (all)    & \cmark & \cmark & \cmark & \xmark & \xmark & \xmark & 2.4 & 3.1 & 2.6 & 2.5 & 2.3 \\
w/o SFBG          & \cmark & \cmark & \cmark & \xmark & \cmark & \cmark & \underline{3.7} & 3.4 & \underline{3.6} & 3.7 & \underline{4.3} \\
w/o RFBG          & \cmark & \cmark & \cmark & \cmark & \xmark & \cmark & 3.4 & \underline{3.8} & 3.2 & 3.4 & 2.1 \\
w/o TCR           & \cmark & \cmark & \cmark & \cmark & \cmark & \xmark & 3.6 & 3.5 & 3.5 & \underline{3.8} & 4.1 \\
\hline
\rowcolor{gray!10}
\textbf{Full CHAT} & \cmark & \cmark & \cmark & \cmark & \cmark & \cmark & \textbf{4.6} & \textbf{4.8} & \textbf{4.6} & \textbf{4.6} & \textbf{4.8} \\
\hline
\end{tabular}
}
\label{tab:ablation_user_study}
\end{table}

\subsection{Downstream Interactive Head Generation}
\label{subsec:ihg}

\noindent To assess CHAT-AVD-50k's utility for downstream dyadic interaction tasks, we evaluate it as pre-training data for \textbf{Interactive Head Generation (IHG)}, where a model generates appropriate facial reactions conditioned on a conversational partner's audio-visual behaviours. We pre-train PerFRDiff \cite{zhu2024perfrdiff} and ReactDiff \cite{luo2025reactdiff} on CHAT-AVD-50k before fine-tuning on the REACT 2024 \cite{song2024react} training split, then evaluate on its test split under the official protocol. As baselines, we report each model trained on REACT 2024 alone. Since CHAT's IFBR block is itself pre-trained on REACT 2024 (Sec. \ref{subsec:experiment}), CHAT-AVD-50k should be regarded as a large-scale diverse augmentation over a related interaction distribution rather than as a fully out-of-domain source.

\noindent Table \ref{tab:ihg_downstream} shows consistent improvements across both reported metrics. Specifically, CHAT-AVD-50k pre-training increases FRCorr from 37.21 to 40.11 for PerFRDiff and from 24.19 to 26.12 for ReactDiff, while reducing FRDist (a dynamic time warping distance to the ground-truth reaction, lower is better) from 94.72 to 89.45 and from 86.70 to 83.87, respectively. We therefore interpret these gains as evidence of useful scale and diversity augmentation within a related distribution, rather than as strictly independent transfer. The consistent gains for two different reaction models indicate that the synthetic dialogues carry interaction patterns that transfer beyond a single architecture.

\begin{table}[t]
\centering
\caption{\textbf{Downstream IHG.} Pre-training PerFRDiff and ReactDiff on CHAT-AVD-50k before fine-tuning on REACT 2024 \cite{song2024react}. FRCorr is reported as $\times10^{-2}$. Best per model in \textbf{bold}.}
\small
\setlength{\tabcolsep}{6pt}
\begin{tabular}{lcccc}
\toprule
\multirow{2}{*}{\textbf{Pre-training}} & \multicolumn{2}{c}{\textbf{PerFRDiff} \cite{zhu2024perfrdiff}} & \multicolumn{2}{c}{\textbf{ReactDiff} \cite{luo2025reactdiff}} \\
\cmidrule(lr){2-3} \cmidrule(lr){4-5}
 & FRCorr$\uparrow$ & FRDist$\downarrow$ & FRCorr$\uparrow$ & FRDist$\downarrow$ \\
\midrule
None (REACT 2024 only) & 37.21 & 94.72 & 24.19 & 86.70 \\
CHAT-AVD-50k & \textbf{40.11} & \textbf{89.45} & \textbf{26.12} & \textbf{83.87} \\
\bottomrule
\end{tabular}
\label{tab:ihg_downstream}
\end{table}

%%%%%%%%%%%%%%%%%%%%%%%%%%%%%%%%%%%%%%%%%%%%
%%%     Conclusion and Limitations      %%%%
%%%%%%%%%%%%%%%%%%%%%%%%%%%%%%%%%%%%%%%%%%%%

% \vspace{-5mm}
\section{Conclusion}
\label{conclusion}
This paper formally defines the DIADG task and proposes CHAT, a framework for generating diverse human-human DIAD clips from a single textual prompt. Experiments show that CHAT outperforms existing related methods on DIAD clip generation, and that CHAT-AVD-50k provides effective pre-training data for downstream IHG, enabling both PerFRDiff and ReactDiff to outperform their real-data-trained baselines on the REACT benchmark. We view CHAT as an initial step towards scalable DIAD synthesis, and we hope that the DIADG task and the CHAT-AVD-50k dataset support further work on large-scale, demographically diverse dyadic interaction data. We will release CHAT-AVD-50k with provenance metadata and bias auditing for responsible use.

\section{Limitations and Future Work}
\label{sec:discussion}

\noindent \textbf{Inference cost.} The main limitation of CHAT is its generation cost. Each dialogue clip runs a large language model, a text-to-speech model and several video diffusion passes, so synthesis at the scale of CHAT-AVD-50k is slow. Here scalable refers to demographic coverage rather than a low per-clip cost, and distillation or faster samplers are a clear next step.

\noindent \textbf{Evaluation and metrics.} FRCorr and FRDiv, taken from facial reaction generation, score the appropriateness and diversity of facial behaviour, not turn-taking timing or dialogue-level semantics. A metric for dialogue-level interaction quality remains open.

\noindent \textbf{Scope.} CHAT currently generates English dialogues, and the downstream study uses REACT 2024, on which the IFBR block is pre-trained. Extending it to more languages and to independent corpora such as IEMOCAP or NoXi is a natural next step.

% \clearpage  % TODO FINAL: This \clearpage needs to be removed from both review and camera-ready versions.

\section*{Acknowledgements}
Thanks to the Bio-inspired Computing and Machine Learning (BCML) Lab at Heriot-Watt University for early research support. Junhao Song is a first-year PhD student in the Department of Computing at Imperial College London, fully funded by the Hitachi-Imperial Centre, a collaboration between Hitachi Ltd, Hitachi Europe and Imperial College London.

\bibliographystyle{splncs04}
\bibliography{main}
\end{document}